\pdfoutput=1
\documentclass[11pt]{article}

\usepackage[arxiv]{acl}  % MAXEDIT: added an arxiv option which is final with page numbers
\usepackage{times}
\usepackage{latexsym}

\usepackage[T1]{fontenc}
\usepackage[utf8]{inputenc}

\usepackage{microtype}

\usepackage{amsmath}
\usepackage{tabularx}
\usepackage{booktabs}
\usepackage{siunitx}
\usepackage{multicol}
\usepackage{multirow}
\usepackage{colortbl}
\usepackage{hhline}
\usepackage{xspace}
\usepackage{pifont}  % for checkmark

\newcommand{\fone}{F$_\text{1}$}

\newcommand{\dataset}[1]{\ensuremath{\mathcal{D_{\mathrm{#1}}}}}

\newcommand{\squad}{SQuAD}
\newcommand{\squadone}{SQuAD1.1}

\newcommand{\dbidaf}{\dataset{BiDAF}}
\newcommand{\dbert}{\dataset{BERT}}
\newcommand{\droberta}{\dataset{RoBERTa}}
\newcommand{\std}[1]{$_{\text{\thinspace{#1}}}$}

\newcommand{\checkmark}{\ding{51}}
\newcommand{\crossmark}{\ding{55}}

\usepackage{graphicx}
\graphicspath{{images/}}

\title{
Models in the Loop: Aiding Crowdworkers with\\ Generative Annotation Assistants
}

\author{
    \textbf{Max Bartolo$^{\dagger\ddagger}$} \quad 
    \textbf{Tristan Thrush$^{\ddagger}$} \quad 
    \textbf{Sebastian Riedel$^{\dagger\ddagger}$} \quad \\
    \textbf{Pontus Stenetorp$^\dagger$} \quad 
    \textbf{Robin Jia$^{\S\ddagger}$} \quad 
    \textbf{Douwe Kiela$^\ddagger$} \\
  $^\dagger$UCL \quad $^\S$USC \quad $^\ddagger$Facebook AI Research \\
  \texttt{m.bartolo@cs.ucl.ac.uk}
}

\begin{document}
\maketitle
\begin{abstract}
In Dynamic Adversarial Data Collection~(DADC), human annotators are tasked with finding examples that models struggle to predict correctly.
Models trained on DADC-collected training data have been shown to be more robust in adversarial and out-of-domain settings, and are considerably harder for humans to fool.
However, DADC is more time-consuming than traditional data collection and thus more costly per annotated example.
In this work, we examine whether we can maintain the advantages of DADC, without incurring the additional cost.
To that end, we introduce Generative Annotation Assistants~(GAAs), generator-in-the-loop models that provide real-time suggestions that annotators can either approve, modify, or reject entirely.
We collect training datasets in twenty experimental settings and perform a detailed analysis of this approach for the task of extractive question answering (QA) for both standard and adversarial data collection.
We demonstrate that GAAs provide significant efficiency benefits with over a 30\% annotation speed-up, while leading to over a 5x improvement in model fooling rates.
In addition, we find that using GAA-assisted training data leads to higher downstream model performance on a variety of question answering tasks over adversarial data collection.
\end{abstract}

\section{Introduction}

%%%%%%%%%%%%%%%%%%%%%%%%%%%%%%
\begin{figure}[t]
    \centering
    \includegraphics[width=\columnwidth]{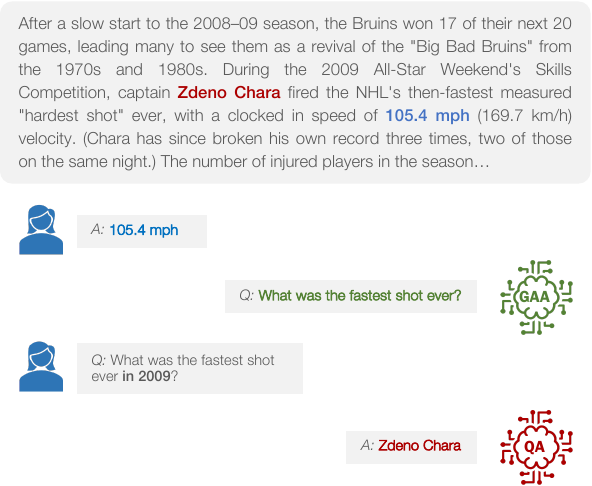}
    \caption{
    Example interaction between an annotator and the models in the loop. The annotator selects an answer from the passage, for which the Generative Annotation Assistant (GAA) prompts a question. The annotator can then freely modify the question and/or answer, or generate another prompt. In the adversarial data collection setting, a model-in-the-loop provides predictions with the aim of encouraging annotators to find model-fooling examples. In the answer prompting setting, an answer suggestion is prompted by the assistive model instead of being selected by the annotator.
    } 
    \label{fig:splash}
\end{figure}
%%%%%%%%%%%%%%%%%%%%%%%%%%%%%%

%
Natural language processing has become increasingly reliant on large datasets obtained using crowd sourcing.
However, crowdsourcing as an unconstrained annotation approach is known to result in machine-exploitable annotator artefacts~\cite{jia2017adversarial,schwartz2017effect,gururangan2018annotation, geva-etal-2019-modeling}, leading to poor out-of-distribution generalisation~\cite{chen-etal-2016-thorough, weissenborn-etal-2017-making, Yogatama2019LearningAE, mccoy-etal-2019-right}.
Dynamic Adversarial Data Collection~(DADC) aims to address these issues by introducing state-of-the-art models into the data collection loop and asking human annotators to produce examples that these models find challenging~\cite{kiela2021dynabench}.
The intuition behind this approach is that it leads human annotators to better explore the space of possible examples.
Previous work has found that DADC leads to improved model robustness on adversarial datasets~\cite{nie-etal-2020-adversarial,bartolo2020beat}, increased sample diversity~\cite{bartolo2020beat,wallace2021analyzing}, better training data \cite{wallace2021analyzing} and better domain generalisation~\cite{bartolo2021improving}.

Despite these advantages, a downside to DADC is that it increases the human effort necessary to annotate a single example and thus the overall annotation cost.
In fact, to date, only a limited number of large-scale training datasets have been produced using DADC and its application has been primarily restricted to producing challenge sets or as additional training data to improve the performance of models already trained on non-DADC curated datasets.
To make better use of DADC data, \citet{bartolo2021improving} propose generating synthetic adversarial training sets to further improve model robustness.
However, this approach inevitably limits example diversity as it relies on examples ultimately generated by a model with no additional human input, and provides no guarantees that useful synthetic examples would transfer across target adversary models of varying capabilities or across annotation rounds.
In this work, we propose assisting annotators by having generative models aid human annotators in the data collection loop.
Concretely, we utilise a Generative Annotation Assistant~(GAA) model that provides prompt suggestions to crowdworkers, while allowing full flexibility for edits and rewrites to support example generation while still allowing for human creativity as shown in Figure~\ref{fig:splash}.
We explore GAAs in a broad range of experimental settings, including standard and adversarial data collection approaches, training on various source datasets, and employing sampling methodologies based on likelihood, adversarial feedback, and uncertainty.
We showcase the value of this approach on the task of extractive question answering~(QA), and find that GAAs can help improve both the standard and adversarial data collection paradigms.
We find considerable efficiency gains, with over 30\% observed annotation speed-ups, as well as improved data effectiveness with up to a 6.1\fone{} improvement in downstream performance over adversarial data collection.
%

%%%%%%%%%%%%%%%%%%%%%%%%%%%%%%
\begin{figure*}[t]
\includegraphics[width=\textwidth]{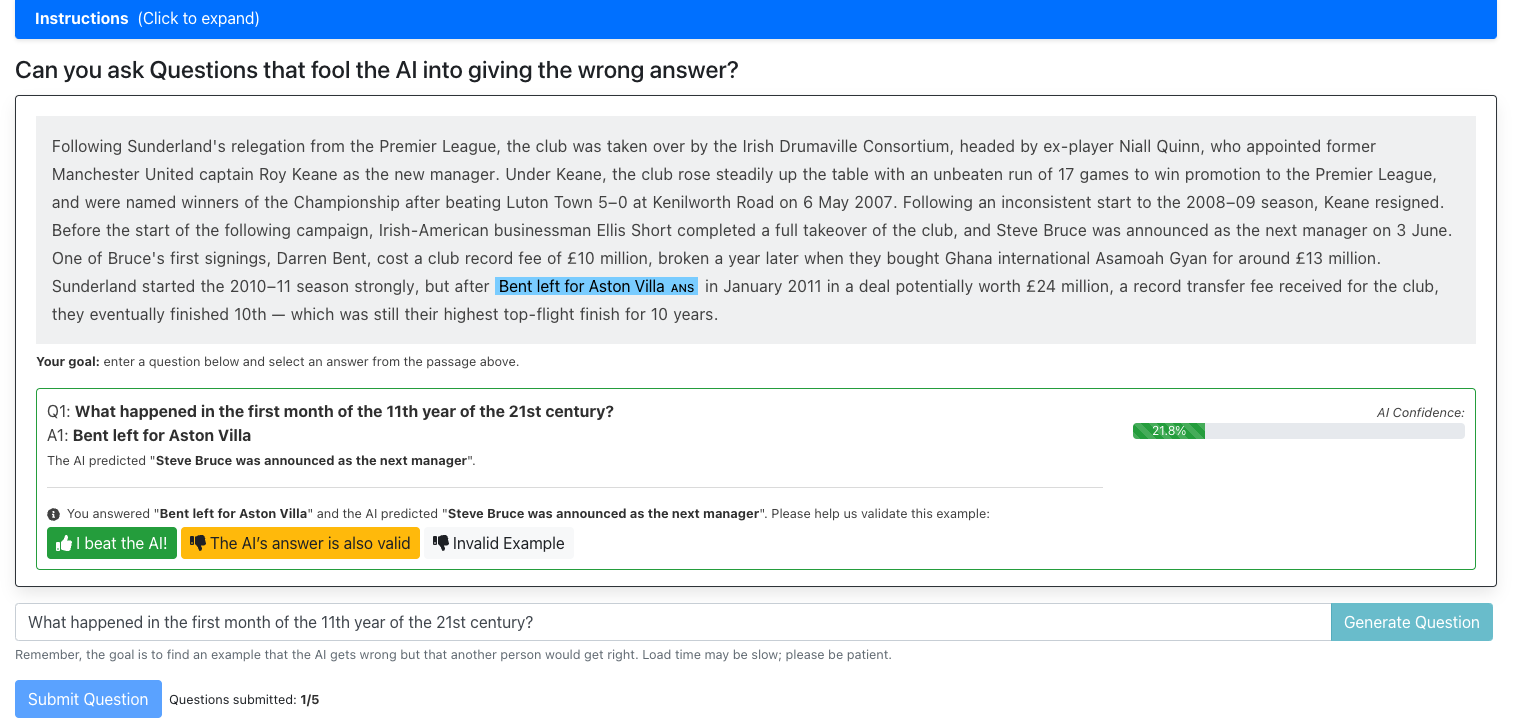}
\caption{
The Annotation Interface used for data collection. This example shows a question generated using a generative assistant trained on the AdversarialQA data and selected an adversarial sampler, which successfully allowed the annotator to beat the QA model in the loop.
}
\label{fig:interface_gaas}
\end{figure*}
%%%%%%%%%%%%%%%%%%%%%%%%%%%%%%

%%%%%%%%%%%%%%%%%%%%%%%%%%%%%%
\section{Related Work}

\subsection{Dynamic Adversarial Data Collection~(DADC)}
There is a rich body of recent work showing the benefits of dynamic adversarial data collection in model evaluation \cite{yang2017mastering,dua2019drop,dinan-etal-2019-build,nie-etal-2020-adversarial,bartolo2020beat,kiela2021dynabench,wallace2021analyzing}, although the approach has been challenged for not necessarily leading to better generalisation on non-adversarial test sets \cite{kaushik-etal-2021-efficacy} and being sensitive to the choice of model that was used in the loop~\cite{bowman2021will,phang2021adversarially}.
This work builds on previous work in adversarial data collection methods for QA~\cite{bartolo2020beat}, and work investigating the use of question generation models to create synthetic adversarial data to improve QA model robustness~\cite{bartolo2021improving}.

\subsection{Generative Model Annotation Support}
A long line of prior work has trained generative models for question answering \citep{du-etal-2017-learning,du-cardie-2018-harvesting,zhao-etal-2018-paragraph,lewis2018generative,alberti-etal-2019-synthetic,puri-etal-2020-training,yang-etal-2020-generative,bartolo2021improving,lewis2021paq}.
In many cases, these approaches filter out questions that an external QA model gets wrong, in order to ensure correctness of the generated questions; our filtering strategies instead focus on generated questions that QA models get wrong as we hypothesise that these would serve as more useful initial prompts to human annotators.
Generative models have also been used to aid experts with writing contrast sets~\citep{wu-etal-2021-polyjuice,ross2021tailor}, but to the best of our knowledge, this is the first work to investigate the use of generative annotation assistants for crowdworkers directly in the annotation loop for NLP.
Recent work on supporting crowdworkers for textual entailment in a non-adversarial setting shows no improvements on downstream transfer performance over baseline, albeit with reductions in previously observed issues with annotation artefacts~\cite{bowman-etal-2020-new}.
Subsequent work highlights the need for further data collection efforts focusing on improving writing-based annotation processes~\cite{vania-etal-2021-comparing}, which we aim to investigate in this work.
Separately,~\citet{ettinger2017buildit} provide \textit{breakers} with the ability to minimally edit original data to identify the boundaries of system capabilities, while~\citet{potts-etal-2020-dynasent} analyse the use of prompts to assist crowdworkers in beating a model in the loop for sentiment analysis.
In both cases, prompts are sourced from existing datasets and are not generated on the fly.
\subsection{Active Learning and Weak Supervision}
Active learning approaches have been used to accelerate annotation~\cite{tsuruoka-etal-2008-accelerating}, although this typically assumes access to a pool or stream of unlabelled data for which the learning algorithm can query labels \citep{settles2009active}.
In our setting, no unlabelled questions are provided, necessitating the use of a generative model to suggest questions instead.
Moreover, our annotators are free to edit and browse generated questions, whereas annotators in active learning typically only provide labels and have no choice in what to label.
Some of our sampling and filtering strategies based on entropy are inspired by uncertainty sampling, a standard active learning algorithm~\citep{lewis1994sequential}.
%

%%%%%%%%%%%%%%%%%%%%%%%%%%%%%%
\section{Experimental Setup}
Our study focuses on the effects of incorporating generative annotation assistants, and understanding their interactions with annotators and discriminative models-in-the-loop in a DADC context for QA.
We provide crowdworkers with a short passage from Wikipedia and ask them to write five questions and highlight the span in the passage that best answers the question for each (see Figure~\ref{fig:interface_gaas}).
We pay workers equally across experiment modes to avoid creating an incentive imbalance and pay out an additional bonus for each question that successfully beats the discriminative QA model i.e., for each question that the model fails to answer correctly.
Finally, we validate all collected examples using a separate worker pool that also undergoes rigorous onboarding and validation. We ask three of these additional workers to report on the validity of each annotated example.
\paragraph{Selected Passages}{
We select passages from KILT~\citep{petroni-etal-2021-kilt} to allow for the possibility of future investigation into cross-domain and task transfer.
We restrict KILT passages to those with between 100 and 600 tokens that are used by at least 5 of the KILT tasks.
Furthermore, we filter out any passages with any 8-gram overlap (after normalisation) to the \squadone{} training or development sets, seeking to ensure that all passages used in our study are novel and previously unseen by the discriminative QA models in the loop.
This leaves a total of 10,109 passages from 421 Wikipedia pages.
We retain and supply all passage-relevant KILT metadata (such as IDs and provenances) with our collected datasets to facilitate future work.
}

\paragraph{Model-in-the-Loop}{
The discriminative QA model in the loop is ELECTRA\textsubscript{Large}~\citep{Clark2020ELECTRA} trained on \squadone{} and AdversarialQA, and enhanced using SynQA to improve adversarial robustness as investigated by~\citet{bartolo2021improving}.\footnote{You can interact with this model at \url{https://dynabench.org/models/109}.}
This model represents the best-performing model on the Dynabench~\cite{kiela2021dynabench} leaderboard at the time of conducting this study, obtaining a word-overlap \fone{} score of 94.5\% on the \squadone{} dev set, and represents the state-of-the-art on AdversarialQA achieving 77.6\% on the \dbidaf{} subset, 71.5\% on \dbert{}, and 63.2\% on \droberta{}.
}

\paragraph{Generator-in-the-Loop}{
For our generative models, we use the \textit{fairseq}~\citep{ott2019fairseq} implementation of BART$_\text{Large}$~\cite{lewis-etal-2020-bart}, and fine-tune the decoder to generate questions conditioned on the passage and the answer highlighted by the annotator.
To provide annotators with a diverse set of questions, we decode using nucleus sampling with $top_p  = 0.75$, as decoding using standard beam search results in questions which are more similar to each other and therefore likely to be less useful as question prompts to annotators.
To speed up inference and model-annotator interaction, we preemptively identify answer candidates for each passage and generate questions to build up a large cache from which we serve questions during annotation.
Once there are no questions remaining in the cache for a particular answer, or if the annotator selects an answer that is not in the cache, we fall back to querying the generative model in real-time. 
In this work, we investigate generative assistants trained on three different sources of questions: \squadone{}, AdversarialQA, and the combination of both \squad{} and AdversarialQA.
}

\paragraph{Question Sampling}{
We investigate three different selection strategies for presenting the generated questions as prompts to annotators: 
i) \textit{generator likelihood} samples candidates in the order prescribed by the generative model's associated likelihood values; 
ii) \textit{adversarial sampling} selects generated questions in order of the least word-overlap \fone{} scores when queried against the discriminative QA model; and 
iii) \textit{uncertainty sampling} is inspired by active learning and selects generated questions in order of the least span selection confidence when queried against the QA model.
The latter two provide an interesting trade-off for exploration as we would expect the quality of the generated questions to be worse than if sampled based on likelihood.
However, we hope that such prompts could serve to inspire annotators and provide a ``starting point'' beyond the answering capabilities of the QA model, irrespective of correctness.
We hypothesise that modifying such examples might be a more effective process for annotators to undertake than when starting from higher quality but less model-confusing prompts, and investigate this question thoroughly.
}

\paragraph{Answer Prompts}{
We also investigate the effects of abstracting away the answer selection task from the annotator.
To identify potential candidate answers, we use Self-Attention Labelling (SAL)~\cite{bartolo2021improving} and investigate providing annotators with both answer prompts as well as the corresponding generated questions.
}

\paragraph{Experimental Settings}{
In total, there are twenty different experimental settings involving combinations of the above-mentioned pipeline components.
We collect 2,000 validated training examples for each of these settings, for a total of 40,000 examples.
For downstream evaluation we train ELECTRA\textsubscript{Large} QA models on the training datasets collected for each setting, and perform identical model selection and hyper-parameter tuning.
}

\paragraph{Annotation Interface}{
We use an adaptation of the Dynabench~\citep{kiela2021dynabench} QA interface that allows annotators to interact with the models in the loop, and further allows them to edit and modify generated questions and answers as required.
The same base interface is used across experimental settings and only varied minimally depending on the current setting, for example by changing the title and instructions in the adversarial annotation setting, or by adding a ``Generate Question'' button when the setting involves GAAs.
In the GAA settings, annotators are not informed what generative model they are interacting with, or what sampling mechanism is being used.
}

%
%%%%%%%%%%%%%%%%%%%%%%%%%%%%%%%%%%%%%%%%%%%%%%%%
\begin{table*}[t]
    \aboverulesep=0pt
    \belowrulesep=0pt
    \renewcommand{\arraystretch}{1.2}
    \centering
    \footnotesize
    \setlength{\tabcolsep}{5.3pt}
        \begin{tabular} {@{\extracolsep{0pt}} c | c c c | c c c c | c @{}}
                \textbf{Adversary-in-the-loop?} & 
                
                \textbf{t (s)} & 
                \textbf{vMER (\%)} & 
                \textbf{t/vMFE (s)} & 
                
                \textbf{\squad{}\textsubscript{dev}} & 
                \textbf{\dataset{BiDAF}} & 
                \textbf{\dataset{BERT}} & 
                \textbf{\dataset{RoBERTa}} & 
                
                \textbf{MRQA} \\
        
        \toprule
            % Exp ID 0
            \crossmark & \textbf{57.2}\std{23.9} & 0.63 & 11,537 & \textbf{85.7} & 43.5 & 28.3 & 21.1 & 52.0 \\ 
            % Exp ID 4
            \checkmark & 61.5\std{27.1} & \textbf{1.86} & \textbf{4,863} & 85.0 & \textbf{53.0} & \textbf{34.2} & \textbf{26.9} & \textbf{58.8} \\
        \bottomrule
        \end{tabular}
    \caption{
    Baseline results comparing standard and adversarial data collection. \textit{t} shows the median time taken per example in seconds and median absolute deviation (subscript). \textit{vMER} is the validated model error rate. \textit{t/vMFE} is the time per validated model-fooling example. Lower is better for the time-dependent metrics. Downstream evaluation is measured by training an ELECTRA\textsubscript{Large} QA model on the collected datasets and evaluating \fone{} scores on the \squadone{} dev set, the AdversarialQA test sets, and the MRQA dev sets for domain generalisation.
    }
    \label{tab:results_baselines}
\end{table*}
%%%%%%%%%%%%%%%%%%%%%%%%%%%%%%%%%%%%%%%%%%%%%%%%
%

\paragraph{Crowdsourcing Protocol}{
We use Amazon Mechanical Turk to recruit workers for this study and run all experiments using Mephisto.\footnote{\url{github.com/facebookresearch/Mephisto}}
To ensure proficiency in English, crowdworkers are required to be based in Canada, the UK, or the US.
They are also required to have a Human Intelligence Task~(HIT) Approval Rate greater than $98\%$, have previously completed at least 1,000 HITs, and undergo a dedicated onboarding process.
Workers were randomly assigned to one of the possible experiment modes and were all presented with passages sampled from the same set, for which they were tasked with writing and answering five questions.
All collected questions were than validated for correctness by a separate group of crowdworkers.
We collect three validations per question and use this information, along with manual verification of a subset of the annotated examples, to maintain a high level of quality and remove examples from workers with less than an 80\% validity rate.
We calculate the reliability of agreement between validators using Fleiss' kappa at 0.46.
Workers were provided an additional bonus for each example validated as having successfully fooled the model in the adversarial data collection settings.
In total, 1,931 workers participated in the study, with 1,559 contributing to the final datasets.
We also continuously validate both annotators and validators based on signals such as repetitiveness, agreement, and manual checks.
}

\paragraph{Evaluation}
We evaluate the outcomes in each of the experimental settings by a selection of metrics: 
\renewcommand{\theenumi}{\roman{enumi}}
\begin{enumerate}
    \item \textit{median time per example} as a measure of annotation efficiency and where a lower time taken is better;
    \item \textit{validated Model Error Rate~(vMER)}~\citep{bartolo2021improving} which evaluates the effectiveness of annotators at generating valid question-answer pairs that the QA model in the loop fails to answer correctly;
    \item \textit{median time per validated model-fooling example} which serves as a single metric incorporating both method efficiency and effectiveness and thus provides a convenient metric for comparison across the various experimental settings; and
    \item \textit{downstream effectiveness} in which we evaluate the performance (by word-overlap \fone{} score) of a QA model trained on the data collected in each of the experimental modes on the standard \squadone{} benchmark, on the AdversarialQA benchmark, and in terms of domain generalisation ability on the MRQA~\citep{fisch2019mrqa} dev sets.
\end{enumerate}
Lower values are better for the time-dependent metrics, however, from the perspective of training data we consider a higher vMER to be better guided by the performance benefits observed for adversarial over standard data collection. This is corroborated by comparison with downstream results.
%

%
%%%%%%%%%%%%%%%%%%%%%%%%%%%%%%%%%%%%%%%%%%%%%%%%
\begin{table*}[t]
    \aboverulesep=0pt
    \belowrulesep=0pt
    \renewcommand{\arraystretch}{1.2}
    \centering
    \footnotesize
    \setlength{\tabcolsep}{4.4pt}
        \begin{tabular} {@{\extracolsep{0pt}} p{0.23\textwidth} | c c c | c c c c | c @{}}
                \textbf{Sampling Strategy} &
                
                \textbf{t (s)} & 
                \textbf{vMER (\%)} & 
                \textbf{t/vMFE (s)} & 
                
                \textbf{\squad{}\textsubscript{dev}} & 
                \textbf{\dataset{BiDAF}} & 
                \textbf{\dataset{BERT}} & 
                \textbf{\dataset{RoBERTa}} & 
                
                \textbf{MRQA} \\
        
        \toprule
            % Exp ID 1
            \textit{Likelihood} & \textbf{37.5}\std{21.4} & 0.62 & 8,708   & \textbf{85.5} & 41.8 & 25.3 & 20.3 & 53.6 \\ 
            % Exp ID 2
            \textit{Adversarial} & 55.6\std{20.9} & \textbf{4.02} & \textbf{1,760}  & 84.7 & \textbf{45.5} & 26.1 & 20.0 & \textbf{54.3} \\ 
            % Exp ID 3
            \textit{Uncertainty} & 63.1\std{28.6} & 2.77 & 3,018  & 83.2 & \textbf{45.5} & \textbf{28.2} & \textbf{21.9} & 53.2 \\ 

        \bottomrule
        \end{tabular}
    \caption{
    Results for the investigation into supporting standard data collection using GAAs. Since this setting assumes no access to adversarially-sourced data, we use a generative model trained only on questions from \squadone{}. There is no adversarial QA model in the loop in this setting.
    }
    \label{tab:results_improving_standard}
\end{table*}
%%%%%%%%%%%%%%%%%%%%%%%%%%%%%%%%%%%%%%%%%%%%%%%%
%

%%%%%%%%%%%%%%%%%%%%%%%%%%%%%%
\section{Results}
Our study allows us to perform a thorough investigation into both the efficiency and effectiveness of the different data annotation methodologies. 
It also allows us to build on work investigating the various differences between standard and adversarial data collection~\cite{kaushik-etal-2021-efficacy}.

\subsection{Standard versus Adversarial Data Collection}
The standard and adversarial data collection settings we use as baselines do not make use of GAAs, and are designed to replicate the \squadone{}~\cite{rajpurkar2016squad} and AdversarialQA~\cite{bartolo2020beat} annotation setups as closely as possible.
However, in contrast to AdversarialQA, our setting only provides annotators with a financial incentive to \emph{try} to beat the model in the loop through the use of a bonus, and does not restrict annotators to only submitting model-fooling examples.

The results, shown in Table~\ref{tab:results_baselines}, highlight the differences between the two annotation approaches.
As expected, standard data collection is more efficient in terms of the time taken per example, as there is no requirement for annotators to make any effort to try to beat a model.
However, the efficiency differences are not as large as seen in settings where annotators \textit{have} to submit model-fooling examples~\cite{bartolo2020beat}.
We also find considerable benefits from adversarial data collection in terms of the validated model error rate and subsequent downstream performance.
As observed by~\citet{bartolo2020beat}, adversarial data collection is more effective on adversarial test sets and aids domain generalisation, with slight performance degradation in the standard evaluation setting, which may be mitigated by increasing the amount of training data or combining with non-adversarial training data (for detailed results, refer to Appendix~\ref{appendix:plussquad}).
The combined performance across evaluation settings is considerably higher for adversarial data collection.
We note that the training data sizes in both these experimental settings are relatively small, and the benefits of adversarial data collection have been shown to be more pronounced in the low data regime, likely due to increased example diversity.
Furthermore, while the passages used in this study are sourced from Wikipedia, there may exist characteristic differences between these and the passages used in \squad{}.
We also observe considerably lower (i.e., better) adversarial human evaluation vMER scores achieved for our synthetically-augmented ELECTRA\textsubscript{Large} model-in-the-loop compared to the 8.8\% reported for RoBERTa\textsubscript{Large} by~\citet{bartolo2021improving}.
We hypothesise that this is primarily due to two factors: the improved robustness of ELECTRA in comparison to RoBERTa, and more tightly-controlled example validation.
For further evidence of the improved adversarial robustness of ELECTRA, refer to Appendix~\ref{appendix:addsent}.
%

%
%%%%%%%%%%%%%%%%%%%%%%%%%%%%%%%%%%%%%%%%%%%%%%%%
\begin{table*}[t]
    \aboverulesep=0pt
    \belowrulesep=0pt
    \renewcommand{\arraystretch}{1.2}
    \centering
    \footnotesize
    \setlength{\tabcolsep}{3.4pt}
        \begin{tabular} {@{\extracolsep{0pt}} p{0.14\textwidth} p{0.11\textwidth} | c c c | c c c c | c @{}}
                \textbf{GAA Training} &
                \textbf{Sampling} &
                
                \textbf{t (s)} & 
                \textbf{vMER (\%)} & 
                \textbf{t/vMFE (s)} & 
                
                \textbf{\squad{}\textsubscript{dev}} & 
                \textbf{\dataset{BiDAF}} & 
                \textbf{\dataset{BERT}} & 
                \textbf{\dataset{RoBERTa}} & 
                
                \textbf{MRQA} \\
        
        \toprule

            % Exp ID 5
            \squad{} & \textit{Likelihood} & 61.1\std{31.9} & 2.25 & 3,501   & \textbf{86.5} & 50.1 & 30.1 & 24.1 & 57.6 \\
            % Exp ID 8
            \squad{} & \textit{Adversarial} & 62.6\std{22.7} & 4.66 & 1,750   & 83.2 & 48.1 & 27.7 & 24.2 & 55.0 \\
            % Exp ID 11
            \squad{} & \textit{Uncertainty} & 62.1\std{26.3} & 2.41 & 3,317   & 86.1 & \textbf{51.8} & 31.1 & 24.4 & \textbf{58.4} \\
            
        \midrule
        
            % Exp ID 6
            AdversarialQA & \textit{Likelihood} & 54.2\std{24.4} & 3.09 & 2,458   & 84.7 & 49.8 & 36.9 & \textbf{29.8} & 56.8 \\
            % Exp ID 9
            AdversarialQA & \textit{Adversarial} & 63.9\std{25.6} & \textbf{6.30} & \textbf{1,262}   & 83.3 & 49.4 & 34.9 & 28.1 & 56.7 \\
            % Exp ID 12
            AdversarialQA & \textit{Uncertainty} & 72.1\std{30.9} & 5.20 & 1,776   & 83.6 & 50.3 & \textbf{37.3} & 27.0 & 55.7 \\
            
        \midrule
        
            % Exp ID 7
            Combined & \textit{Likelihood} & \textbf{53.6}\std{25.6} & 2.64 & 2,724   & 85.1 & 48.9 & 33.4 & 24.7 & 56.5 \\
            % Exp ID 10
            Combined & \textit{Adversarial} & 69.6\std{29.6} & 4.86 & 1,922   & 82.9 & 48.0 & 33.6 & 28.6 & 54.5 \\
            % Exp ID 13
            Combined & \textit{Uncertainty} & 62.0\std{25.0} & 4.87 & 1,690   & 85.3 & 50.5 & 33.9 & 28.8 & 56.5 \\
            
        \bottomrule
        \end{tabular}
    \caption{
    Results for the investigation into supporting adversarial data collection using GAAs. We investigate three different GAA training dataset sources, and three sampling strategies. The adversarial QA model used in the annotation loop is identical for all settings.
    }
    \label{tab:results_improving_adversarial}
\end{table*}
%%%%%%%%%%%%%%%%%%%%%%%%%%%%%%%%%%%%%%%%%%%%%%%%
%

%
\subsection{Improving Standard Data Collection}
We now investigate whether it might be possible to improve standard data collection practices using generative assistants -- \textit{can we achieve similar performance to adversarial data collection without access to any adversarial data?}
We therefore use a GAA trained on \squadone{}, and investigate the three sampling techniques namely: likelihood, adversarial, and uncertainty sampling.
Results are shown in Table~\ref{tab:results_improving_standard}.
We find that using a GAA with likelihood sampling considerably improves the efficiency of the annotation process in comparison to the standard data collection baseline in Table~\ref{tab:results_baselines}.
It also gives comparable vMER results and downstream QA performance.
Furthermore, both the adversarial and uncertainty sampling strategies prove effective.
While the reduction in time taken per example is not as substantial as for standard likelihood sampling, and is comparable to the standard data collection baseline, the vMER -- an indicator of the diversity of the collected training data -- is substantially improved and outperforms the adversarial data collection baseline.
The downstream results are also promising, providing slight improvements over the standard data collection setting, particularly with regards to domain generalisation.
They start to make progress towards the values obtained for the adversarial data collection baseline although, despite the improved vMER, overall downstream performance is considerably higher in the adversarial data collection setting.
In summary, these results shows that we can encourage annotators to come up with more challenging examples without requiring any adversarially-collected data or an adversarial model in the loop, simply through the use of GAAs paired with an appropriate sampling strategy.
However, using adversarial data collection still provides substantially better downstream performance.
These observations are in line with our initial hypothesis that sampling generated prompts from regions of known model uncertainty, or prompts that we know the model finds challenging to answer, irrespective of generated sample quality, provides annotators with a better starting point for example creation.
%

%
%%%%%%%%%%%%%%%%%%%%%%%%%%%%%%%%%%%%%%%%%%%%%%%%
\begin{table*}[t]
    \aboverulesep=0pt
    \belowrulesep=0pt
    \renewcommand{\arraystretch}{1.2}
    \centering
    \footnotesize
    \setlength{\tabcolsep}{3.4pt}
        \begin{tabular} {@{\extracolsep{0pt}} p{0.14\textwidth} p{0.11\textwidth} | c c c | c c c c | c @{}}
                \textbf{GAA Training} &
                \textbf{Sampling} &
                
                \textbf{t (s)} & 
                \textbf{vMER (\%)} & 
                \textbf{t/vMFE (s)} & 
                
                \textbf{\squad{}\textsubscript{dev}} & 
                \textbf{\dataset{BiDAF}} & 
                \textbf{\dataset{BERT}} & 
                \textbf{\dataset{RoBERTa}} & 
                
                \textbf{MRQA} \\
        
        \toprule

            % Exp ID 15
            AdversarialQA & \textit{Likelihood} & \textbf{38.5}\std{22.7} & 5.63 & 988          & 83.8 & 49.7 & \textbf{40.3} & \textbf{30.7} & 55.9 \\
            % Exp ID 17
            AdversarialQA & \textit{Adversarial} & 44.2\std{21.6} & 9.46 & \textbf{668} & 83.9 & 48.7 & 36.2 & 30.3 & 55.2 \\
            % Exp ID 19
            AdversarialQA & \textit{Uncertainty} & 49.9\std{24.8} & 7.80 & 854         & 84.8 & \textbf{51.3} & 38.9 & 30.6 & 56.3 \\
            
        \midrule
        
            % Exp ID 14
            Combined & \textit{Likelihood} & 45.9\std{23.4} & 2.90 & 2,196               & \textbf{85.2} & 51.1 & 37.5 & 28.1 & \textbf{56.4} \\
            % Exp ID 16
            Combined & \textit{Adversarial} & 56.3\std{27.1} & \textbf{9.53} & 785              & 83.5 & 48.4 & 35.5 & 29.3 & 55.5 \\
            % Exp ID 18
            Combined & \textit{Uncertainty} & 57.6\std{27.7} & 4.48 & 1,841              & 83.4 & 47.4 & 36.6 & 27.8 & 55.2 \\
            
        \bottomrule
        \end{tabular}
    \caption{
    Results for the investigation into supporting adversarial data collection using GAAs equipped with answer prompting. We investigate two different GAA training dataset sources, and three sampling strategies. The adversarial QA model-in-the-loop is identical for all settings.
    }
    \label{tab:results_answer_prompting}
\end{table*}
%%%%%%%%%%%%%%%%%%%%%%%%%%%%%%%%%%%%%%%%%%%%%%%%
%

%
\subsection{Improving Adversarial Data Collection}
Following the efficiency gains observed for standard data collection, we investigate whether it is possible for GAAs to provide further improvements over adversarial data collection.
As for the previous experiments, we investigate GAAs trained on three different datasets: \squadone{}, AdversarialQA, and the combination of both.
We combine each of these with the three previously discussed sampling strategies resulting in nine different experimental settings.
Results are shown in Table~\ref{tab:results_improving_adversarial}.
We find that when annotators are incentivised to try to beat an adversarial QA model-in-the-loop, the previously seen efficiency gains are not as clear cut.
In fact, annotators are slightly slower than for the adversarial data collection baseline when using a \squad{}-trained GAA.
When using a GAA that has been trained on adversarially-sourced questions, standard likelihood sampling provides efficiency gains over the baseline, however, both adversarial and uncertainty sampling (which naturally lead to more complex prompts that might be more challenging to work with) actually slow annotators down, although they do provide improved validated model error rates and overall better adversarial example generation efficiency measured by the time taken per validated model-fooling example.
In terms of downstream performance, there is no clear best option, but the best settings consistently outperform the adversarial data collection baseline on the most challenging examples (\dbert{} and \droberta{}) while providing comparable results in the other evaluation settings.
Surprisingly, we find that various settings, particularly those involving a \squad{}-trained GAA can provide performance gains over the standard data collection baseline on \squadone{}.
We also observe that a \squad{}-trained GAA with uncertainty sampling gives best performance on the less challenging evaluation sets, while an AdversarialQA-trained GAA gives best performance on the evaluation datasets collected using a more performant adversary.
This is also in line with the observations made by~\citet{bartolo2020beat} showing a distributional shift in question type and complexity with an increasingly stronger model-in-the-loop.
The general takeaway in terms of the ideal experimental setting from the perspective of downstream performance is that it depends on the particular evaluation setting, with GAAs trained on examples from a particular setting yielding better performance when the downstream model is also evaluated in similar conditions.
Another key observation is that both the validated model error rate and time per validated model-fooling example comfortably outperform the baselines across the board, highlighting the enhancements to the effectiveness of the annotation process provided by incorporating GAAs in the loop.

\subsection{Investigating Answer Prompting}
The settings explored in the previous sections focus on investigating the effects of assisting free-text generation of the questions using GAAs.
However, the QA crowdsourcing setting also involves annotation of answer spans, which we also explore in the search for efficiency gains.
Here, we explore GAAs trained on datasets with adversarially-sourced components and the same three sampling strategies as previously (likelihood, uncertainty and adversarial), while additionally providing annotators with an answer suggestion.
In essence, this is similar to an answer and question validation setting, with the difference that annotators have the ability to freely modify both answer and question, or request additional suggestions.
Results for our experiments involving answer assistance are shown in Table~\ref{tab:results_answer_prompting}.
We find that answer prompting is very effective at improving annotation efficiency, providing gains in all six experimental settings while also providing improved vMER results in all cases.
We also see very similar downstream performance result patterns to the previous set of experiments -- for performance on the more challenging evaluation sets (\dbert{} and \droberta{}), an AdversarialQA-trained GAA with likelihood sampling gives best performance, while for performance on \squad{}, a GAA trained on examples including \squad{} gives the best results.
As previously discussed and as shown in Appendix~{\ref{appendix:plussquad}}, adding \squad{} examples to the training data mitigates this effect. 
The consistency in performance patterns serves to further highlight the previous observation that, while using GAAs provides considerable gains in both the efficiency of the annotation process and effectiveness in terms of downstream results, the ideal annotation setup should be selected based on the target downstream evaluation.
It is also worth highlighting the considerable performance improvements on the more challenging AdversarialQA evaluation sets observed when using an AdversarialQA-trained GAA even over adversarial data collection.
%

%%%%%%%%%%%%%%%%%%%%%%%%%%%%%%
\section{Annotator Interaction with GAAs}
While we provide annotators with instructions explaining how they can use the GAAs to aid their annotation, they are free to query the generative models as many times as they like, if at all, during annotation.
We are interested to see how the three main factors affecting interaction with the GAAs that we explore -- training data, sampling strategy, and answer prompting -- affect the ways in which annotators interact or use the GAAs.
Results, shown in Table~\ref{tab:qs_per_example}, indicate that annotators query the GAA less frequently when being shown simpler prompts i.e. those obtained using a GAA trained on non-adversarially sourced examples, or selected using likelihood sampling which tends to provide higher quality and less complex generated texts.
We also find that annotators query the GAA more frequently when an answer prompt is also provided.
We believe that this can be attributed to the fact that the answer and question prompt setting is more similar to a validation workflow, allowing annotators to generate prompts until a satisfactory one is found.
\begin{table}
\centering
\footnotesize
\setlength{\tabcolsep}{5.5pt}
\begin{tabular}{ l c c }
    \multirow{2}{*}{\textbf{Feature}} & \multirow{2}{*}{\textbf{Setting}} & \multirow{2}{7.9em}{\textbf{Avg. \#Generations per Example}}
    \\
    \\
    
    \midrule
    
    \multirow{3}{*}{GAA Training} & SQuAD & 0.69 \\
    & AdversarialQA & 0.86 \\
    & Combined & 0.83 \\
    
    \midrule
    
    \multirow{3}{*}{Sampling} & \textit{Likelihood} & 0.67 \\
    & \textit{Adversarial} & 0.87 \\
    & \textit{Uncertainty} & 0.83 \\
    
    \midrule

    \multirow{2}{*}{Answer Prompt?} & \crossmark{} & 0.73 \\
    & \checkmark{} & 0.91 \\
    
    \bottomrule
\end{tabular}
\caption{
Results showing how often annotators query the GAA for different experimental settings.
}
\label{tab:qs_per_example}
\end{table}
%

%%%%%%%%%%%%%%%%%%%%%%%%%%%%%%
\section{Discussion and Conclusion}
In this work, we introduce Generative Annotation Assistants (GAAs) and investigate their potential to aid crowdworkers with creating more effective training data more efficiently.
We perform a thorough analysis of how GAAs can be used for improving QA dataset annotation in different settings, including different generative model training data, sampling strategies, and whether to also provide annotators with answer suggestions.
We find that GAAs are beneficial in both the standard and adversarial data collection settings.
In the standard data collection setting, and under the assumption of no access to adversarially-collected data, GAAs with prompts sampled based on likelihood provide annotation speed-ups, while prompts sampled by adversarial performance or uncertainty metrics provide benefits to both the model error rates on the collected data as well as subsequent downstream QA performance.
We find that while GAAs are effective for improving standard data collection, we still do not approach the performance obtained when using adversarial data collection.
For adversarial data collection, we demonstrate improved effectiveness of the annotation process over the non-GAA baseline, although this comes at a cost of reduced annotation efficiency.
We show that also aiding annotators with answer prompts boosts data collection efficiency even beyond that of standard data collection, while retaining overall downstream performance.
We find that the ideal annotation setting differs for different intended evaluations, with an uncertainty-sampled GAA trained on data that was not adversarially-collected providing best performance on simpler questions, while a GAA trained on adversarially-collected data provides best downstream performance on more challenging evaluation sets.
However, we also find that combining with a small sample of \squad{} training examples can boost performance on these less-challenging questions, and that in this setting a likelihood-sampled adversarially-trained GAA consistently provides the best results.
In terms of efficiency, we see annotation speed-ups over baseline of 34.4\% for standard data collection and 37.4\% for adversarial data collection.
In terms of effectiveness, we see over a 5x improvement in vMER over adversarial data collection, along with downstream performance gains.
We improve over standard data collection on \squad{}\textsubscript{dev} by up to 0.8\fone{} and improve over adversarial data collection by up to 6.1\fone{} on \dbert{}, and 3.8\fone{} on \droberta{}.
Furthermore, we see benefits in domain generalisation over standard data collection, and show that annotators interact with the GAA more frequently when it has been trained on adversarially-collected data, is sampled based on adversarial or uncertainty feedback, and also provides answer prompts.

While our analysis is limited by the size of the collected data, we believe that GAAs can help drive further innovation into improved data collection methodologies based on these findings. 
We hope that our analysis of various aspects of GAA incorporation into the annotation pipeline and the interactions between annotators and multiple models in the loop can help inform future work exploring broader aspects of GAA use, such as for other NLP tasks or for larger scale annotation efforts.
%

%%%%%%%%%%%%%%%%%%%%%%%%%%%%%%
%
\section{Ethical Considerations}
We collect a training datasets as a part of the analysis in this work. The passages are sourced from Wikipedia through KILT.
As described in the main text, our incentive structure is designed to ensure that crowdworkers were fairly compensated.
Our datasets focus on the English language. As this data is not collected for the purpose of designing NLP applications, we do not foresee any risks associated with the use of this data.
%
%%%%%%%%%%%%%%%%%%%%%%%%%%%%%%

% %%%%%%%%%%%%%%%%%%%%%%%%%%%%%%
\section*{Acknowledgements}
The authors would like to thank the Dynabench team for their feedback and continuous support.
%

% \clearpage

%%%%%%%%%%%%%%%%%%%%%%%%%%%%%%
\bibliography{anthology,custom}
\bibliographystyle{acl_natbib}

\clearpage

\appendix

%%%%%%%%%%%%%%%%%%%%%%%%%%%%%%
\section{Breakdown of MRQA Results}
\label{appendix:mrqa}

%
%%%%%%%%%%%%%%%%%%%%%%%%%%%%%%%%%%%%%%%%%%%%%%%%
\begin{table*}[t]
    \aboverulesep=2pt
    \belowrulesep=2pt
    \renewcommand{\arraystretch}{1.5}
    \centering
    \footnotesize
    \setlength{\tabcolsep}{2pt}
        \begin{tabular} {@{\extracolsep{0pt}} p{0.19\textwidth} c | c c c c c c c c c c c c @{}}
                \multirow{2}{10em}{\textbf{GAA Details}} &
                \multirow{2}{2.3em}{\textbf{Adv?}} &
                \multicolumn{6}{c}{\textit{MRQA in-domain}} &
                \multicolumn{6}{c}{\textit{MRQA out-of-domain}} \\
                \hhline{~~||------||------}
                & &
                \scriptsize HotpotQA & 
                \scriptsize NQs & 
                \scriptsize NewsQA & 
                \scriptsize SearchQA & 
                \scriptsize \squad{} & 
                \scriptsize TriviaQA & 
                \scriptsize BioASQ & 
                \scriptsize DROP & 
                \scriptsize DuoRC & 
                \scriptsize RACE & 
                \scriptsize RelExt. & 
                \scriptsize TextbookQA \\
        
        \toprule

            % Exp ID 0
            - & \crossmark & 64.5 & 58.9 & 51.7 & 16.5 & 84.9 & 55.9 & 61.5 & 28.1 & 50.4 & 38.1 & 81.9 & 31.3 \\
            
            % Exp ID 4
            - & \checkmark & 65.3 & \textbf{65.4} & \textbf{57.2} & 36.6 & 85.0 & 62.2 & 66.9 & 43.4 & 55.5 & 44.1 & 83.4 & 41.1 \\
            
        \midrule
            
            % Exp ID 1
            \squad{} (\textit{Likelihood}) & \crossmark & 58.0 & 61.5 & 53.9 & 23.3 & 85.4 & 59.5 & 64.0 & 32.5 & 54.2 & 36.3 & 81.6 & 33.4 \\
            % Exp ID 2
            \squad{} (\textit{Adversarial}) & \crossmark & 56.7 & 63.8 & 55.4 & \textbf{37.4} & 84.1 & 55.4 & 60.7 & 30.4 & 53.9 & 38.8 & 75.5 & 39.3 \\
            % Exp ID 3
            \squad{} (\textit{Uncertainty}) & \crossmark & 62.3 & 62.0 & 53.5 & 26.3 & 83.4 & 57.6 & 63.0 & 30.9 & 50.5 & 37.0 & 79.6 & 32.2 \\
            
        \midrule
            % Exp ID 5
            \squad{} (\textit{Likelihood}) & \checkmark & \textbf{67.3} & 62.9 & 56.9 & 30.3 & \textbf{86.7} & 60.8 & 66.1 & 39.0 & 54.7 & 43.2 & 81.8 & 41.8 \\
            % Exp ID 8
            \squad{} (\textit{Adversarial}) & \checkmark & 60.5 & 61.6 & 51.4 & 30.8 & 83.3 & 59.7 & 62.5 & 38.1 & 52.7 & 39.9 & 80.7 & 39.5 \\
            % Exp ID 11
            \squad{} (\textit{Uncertainty}) & \checkmark & 62.7 & 65.3 & 55.9 & 32.9 & 86.3 & 63.8 & 65.2 & 40.4 & 56.4 & \textbf{44.4} & 81.3 & \textbf{46.0} \\
            
        \midrule
        
            % Exp ID 6
            AdvQA (\textit{Likelihood}) & \checkmark & 63.5 & 62.6 & 53.3 & 23.5 & 84.9 & 59.9 & 66.7 & \textbf{49.5} & 53.3 & 41.5 & 83.7 & 39.4 \\
            % Exp ID 9
            AdvQA (\textit{Adversarial}) & \checkmark & 60.8 & 63.2 & 52.9 & 33.2 & 83.8 & 59.5 & 63.9 & 44.8 & 53.4 & 40.8 & 81.4 & 43.2 \\
            % Exp ID 12
            AdvQA (\textit{Uncertainty}) & \checkmark & 62.3 & 61.8 & 53.3 & 26.0 & 83.6 & \textbf{64.0} & 62.5 & 47.6 & 52.9 & 39.4 & 81.8 & 32.9 \\
            
        \midrule
        
            % Exp ID 7
            Combined (\textit{Likelihood}) & \checkmark & 61.6 & 62.6 & 56.1 & 21.3 & 85.0 & 58.8 & \textbf{67.7} & 46.9 & \textbf{56.5} & 43.3 & 79.1 & 39.6 \\
            % Exp ID 10
            Combined (\textit{Adversarial}) & \checkmark & 60.8 & 60.4 & 51.8 & 30.0 & 82.7 & 55.7 & 61.2 & 42.6 & 53.5 & 38.5 & 79.6 & 37.4 \\
            % Exp ID 13
            Combined (\textit{Uncertainty}) & \checkmark & 64.7 & 64.2 & 53.5 & 27.3 & 85.4 & 59.1 & 64.4 & 45.5 & 49.2 & 41.1 & 83.5 & 40.6 \\
            
        \midrule
            & \multicolumn{12}{c}{\textit{Results below are for the settings with answer prompting}} \\
        \midrule
        
            % Exp ID 15
            AdvQA (\textit{Likelihood}) & \checkmark & 60.4 & 63.9 & 51.8 & 26.8 & 83.5 & 56.9 & 65.8 & 48.4 & 51.7 & 42.5 & 81.0 & 38.0 \\
            % Exp ID 17
            AdvQA (\textit{Adversarial}) & \checkmark & 60.0 & 63.8 & 51.3 & 25.0 & 83.7 & 60.4 & 65.0 & 48.6 & 49.9 & 40.4 & 83.3 & 31.0 \\
            % Exp ID 19
            AdvQA (\textit{Uncertainty}) & \checkmark & 62.7 & 64.0 & 51.2 & 32.9 & 84.9 & 58.3 & 66.3 & 47.0 & 45.4 & 42.3 & 83.0 & 37.9 \\
            
        \midrule
        
            % Exp ID 14
            Combined (\textit{Likelihood}) & \checkmark & 63.4 & 63.9 & 55.1 & 24.5 & 83.2 & 60.6 & 66.7 & 47.0 & 55.9 & 39.3 & 82.2 & 34.9 \\
            % Exp ID 16
            Combined (\textit{Adversarial}) & \checkmark & 62.0 & 63.7 & 51.6 & 18.2 & 83.5 & 60.6 & 64.6 & 48.5 & 53.4 & 40.4 & \textbf{83.8} & 35.3 \\
            % Exp ID 18
            Combined (\textit{Uncertainty}) & \checkmark & 60.6 & 62.9 & 54.2 & 25.0 & 83.6 & 59.2 & 63.3 & 44.4 & 52.5 & 41.6 & 80.3 & 35.3 \\
            
        \bottomrule
        \end{tabular}
    \caption{
    Result breakdown for all twenty experiment modes on the MRQA evaluation sets.
    }
    \label{tab:results_mrqa}
\end{table*}
%%%%%%%%%%%%%%%%%%%%%%%%%%%%%%%%%%%%%%%%%%%%%%%%
%

%
Table~\ref{tab:results_mrqa} shows the breakdown of results on the 12 MRQA in- and out- of domain evaluation sets.
%
%%%%%%%%%%%%%%%%%%%%%%%%%%%%%%

%%%%%%%%%%%%%%%%%%%%%%%%%%%%%%
\section{Combining with \squadone{}}
\label{appendix:plussquad}

\citet{bartolo2020beat} and subsequent works find that the performance degradation in the original evaluation setting when training on adversarially-collected data only is mitigated by also including some of the original training data.
To investigate this further, we combine and shuffle the training datasets collected in each of the experimental settings with 2k \squadone{} examples for a total of 4k training examples per experiment.
The baseline results in Table~\ref{tab:results_plussquad_baselines} show that this results in similar performance, if slightly improved on the \squad{} dev set, when using some adversarially-collected data.
We also show the results for the other experimental settings in Tables~\ref{tab:results_plussquad_improving_standard}, \ref{tab:results_plussquad_improving_adversarial} and \ref{tab:results_plussquad_answer_prompting}, noting very similar performance variation between settings as those reported earlier. 
%

%
%%%%%%%%%%%%%%%%%%%%%%%%%%%%%%%%%%%%%%%%%%%%%%%%
\begin{table*}[t]
    \aboverulesep=0pt
    \belowrulesep=0pt
    \renewcommand{\arraystretch}{1.2}
    \centering
    \footnotesize
    \setlength{\tabcolsep}{18.0pt}
        \begin{tabular} {@{\extracolsep{0pt}} c | c c c c | c }
                \textbf{Adversary-in-the-loop?} & 
                
                \textbf{\squad{}\textsubscript{dev}} & 
                \textbf{\dataset{BiDAF}} & 
                \textbf{\dataset{BERT}} & 
                \textbf{\dataset{RoBERTa}} & 
                
                \textbf{MRQA} \\
        
        \toprule
            % Exp ID 0
            \crossmark & 88.9 & 49.8 & 28.6 & 22.7 & 56.1 \\ 
            % Exp ID 4
            \checkmark & \textbf{89.3} & \textbf{53.2} & \textbf{34.1} & \textbf{27.3} & \textbf{60.1} \\
        \bottomrule
        \end{tabular}
    \caption{
    Baseline results comparing standard and adversarial data collection. Downstream evaluation is measured by training an ELECTRA\textsubscript{Large} QA model on each of the collected datasets combined with 2k \squad{} training examples (for a total of 4k examples) and evaluating \fone{} scores on the \squadone{} dev set, the AdversarialQA test sets, and the MRQA dev sets for domain generalisation.
    }
    \label{tab:results_plussquad_baselines}
\end{table*}
%%%%%%%%%%%%%%%%%%%%%%%%%%%%%%%%%%%%%%%%%%%%%%%%
%

%
%%%%%%%%%%%%%%%%%%%%%%%%%%%%%%%%%%%%%%%%%%%%%%%%
\begin{table*}[t]
    \aboverulesep=0pt
    \belowrulesep=0pt
    \renewcommand{\arraystretch}{1.2}
    \centering
    \footnotesize
    \setlength{\tabcolsep}{13.0pt}
        \begin{tabular} {@{\extracolsep{0pt}} p{0.32\textwidth} | c c c c | c }
                \textbf{Sampling Strategy} &
                
                \textbf{\squad{}\textsubscript{dev}} & 
                \textbf{\dataset{BiDAF}} & 
                \textbf{\dataset{BERT}} & 
                \textbf{\dataset{RoBERTa}} & 
                
                \textbf{MRQA} \\
        
        \toprule
            % Exp ID 1
            \textit{Likelihood} & \textbf{88.9} & 49.2 & 29.0 & 22.8 & 56.7 \\ 
            % Exp ID 2
            \textit{Adversarial} & 88.7 & \textbf{52.0} & \textbf{30.1} & \textbf{24.5} & \textbf{58.1} \\ 
            % Exp ID 3
            \textit{Uncertainty} & 88.5 & 50.0 & 29.7 & 22.8 & 57.1 \\ 

        \bottomrule
        \end{tabular}
    \caption{
    Results for the investigation into supporting standard data collection using GAAs when combining with 2k \squad{} training examples. There is no adversarial QA model in the loop in this setting.
    }
    \label{tab:results_plussquad_improving_standard}
\end{table*}
%%%%%%%%%%%%%%%%%%%%%%%%%%%%%%%%%%%%%%%%%%%%%%%%
%

%
%%%%%%%%%%%%%%%%%%%%%%%%%%%%%%%%%%%%%%%%%%%%%%%%
\begin{table*}[t]
    \aboverulesep=0pt
    \belowrulesep=0pt
    \renewcommand{\arraystretch}{1.2}
    \centering
    \footnotesize
    \setlength{\tabcolsep}{13.0pt}
        \begin{tabular} {@{\extracolsep{0pt}} p{0.14\textwidth} p{0.12\textwidth} | c c c c | c}
                \textbf{GAA Training} &
                \textbf{Sampling} &
                
                \textbf{\squad{}\textsubscript{dev}} & 
                \textbf{\dataset{BiDAF}} & 
                \textbf{\dataset{BERT}} & 
                \textbf{\dataset{RoBERTa}} & 
                
                \textbf{MRQA} \\
        
        \toprule

            % Exp ID 5
            \squad{} & \textit{Likelihood} & 89.4 & 51.4 & 31.7 & 24.1 & 58.8 \\
            % Exp ID 8
            \squad{} & \textit{Adversarial} & 88.4 & 50.9 & 31.8 & 23.2 & 59.3 \\
            % Exp ID 11
            \squad{} & \textit{Uncertainty} & \textbf{89.6} & 53.8 & 31.6 & 25.3 & 59.0 \\
            
        \midrule
        
            % Exp ID 6
            AdversarialQA & \textit{Likelihood} & 89.3 & 52.4 & 38.1 & \textbf{30.2} & \textbf{60.6} \\
            % Exp ID 9
            AdversarialQA & \textit{Adversarial} & 88.8 & 54.0 & 34.5 & 27.0 & 59.3 \\
            % Exp ID 12
            AdversarialQA & \textit{Uncertainty} & 88.8 & 54.5 & \textbf{39.2} & 30.0 & 58.3 \\
            
        \midrule
        
            % Exp ID 7
            Combined & \textit{Likelihood} & 88.9 & \textbf{54.6} & 37.4 & 27.7 & 58.7 \\
            % Exp ID 10
            Combined & \textit{Adversarial} & 89.0 & 53.2 & 34.9 & 25.7 & 58.3 \\
            % Exp ID 13
            Combined & \textit{Uncertainty} & 88.9 & 54.4 & 35.9 & 26.9 & 57.8 \\
            
        \bottomrule
        \end{tabular}
    \caption{
    Results for the investigation into supporting adversarial data collection using GAAs when combining with 2k \squad{} training examples. We investigate three different GAA training dataset sources, and three sampling strategies. The adversarial QA model used in the annotation loop is identical for all settings.
    }
    \label{tab:results_plussquad_improving_adversarial}
\end{table*}
%%%%%%%%%%%%%%%%%%%%%%%%%%%%%%%%%%%%%%%%%%%%%%%%
%

%
%%%%%%%%%%%%%%%%%%%%%%%%%%%%%%%%%%%%%%%%%%%%%%%%
\begin{table*}[t]
    \aboverulesep=0pt
    \belowrulesep=0pt
    \renewcommand{\arraystretch}{1.2}
    \centering
    \footnotesize
    \setlength{\tabcolsep}{13.0pt}
        \begin{tabular} {@{\extracolsep{0pt}} p{0.14\textwidth} p{0.12\textwidth} | c c c c | c }
                \textbf{GAA Training} &
                \textbf{Sampling} &
                
                \textbf{\squad{}\textsubscript{dev}} & 
                \textbf{\dataset{BiDAF}} & 
                \textbf{\dataset{BERT}} & 
                \textbf{\dataset{RoBERTa}} & 
                
                \textbf{MRQA} \\
        
        \toprule

            % Exp ID 15
            AdversarialQA & \textit{Likelihood} & \textbf{89.2} & 53.9 & \textbf{43.4} & \textbf{31.9} & 59.7 \\
            % Exp ID 17
            AdversarialQA & \textit{Adversarial} & 89.1 & 53.4 & 36.4 & 28.0 & 58.8 \\
            % Exp ID 19
            AdversarialQA & \textit{Uncertainty} & 88.5 & \textbf{55.4} & 37.6 & 27.5 & 59.0 \\
            
        \midrule
        
            % Exp ID 14
            Combined & \textit{Likelihood} & \textbf{89.2} & 55.0 & 38.4 & 29.8 & \textbf{61.0} \\
            % Exp ID 16
            Combined & \textit{Adversarial} & 88.6 & 54.7 & 37.7 & 29.4 & 59.4 \\
            % Exp ID 18
            Combined & \textit{Uncertainty} & 88.8 & 53.1 & 32.6 & 26.7 & 57.5 \\
            
        \bottomrule
        \end{tabular}
    \caption{
    Results for the investigation into supporting adversarial data collection using GAAs equipped with answer prompting when combining with 2k \squad{} training examples. We investigate two different GAA training dataset sources, and three sampling strategies. The adversarial QA model-in-the-loop is identical for all settings.
    }
    \label{tab:results_plussquad_answer_prompting}
\end{table*}
%%%%%%%%%%%%%%%%%%%%%%%%%%%%%%%%%%%%%%%%%%%%%%%%
%

%%%%%%%%%%%%%%%%%%%%%%%%%%%%%%

%%%%%%%%%%%%%%%%%%%%%%%%%%%%%%
\section{Adversarial Robustness of ELECTRA and RoBERTa}
\label{appendix:addsent}

\begin{table*}
\centering
\setlength{\tabcolsep}{6.4pt}
\begin{tabular}{llrrr}
    \textbf{Model} & \textbf{Training Data} & \textbf{SQuAD\textsubscript{dev}} & \textbf{AddSent} & \textbf{AddOneSent}\\
    \midrule
    
    \multirow{2}{*}{BERT\textsubscript{Large}} & SQuAD & 90.3 & 73.7 & 80.3 \\
    & SQuAD + AdversarialQA & \textbf{93.3} & \textbf{80.1} & \textbf{85.2} \\
    
    \midrule
    
    \multirow{4}{*}{RoBERTa\textsubscript{Large}} & SQuAD & 93.5 & 82.4 & 86.9 \\
    & SQuAD + AdversarialQA & 92.5 & 83.4 & 86.7 \\
    & SQuAD + AdversarialQA + SynQA & 94.8 & 86.0 & 89.0 \\
    & SQuAD + AdversarialQA + SynQA\textsubscript{Ext} & \textbf{94.9} & \textbf{87.1} & \textbf{90.1} \\
    
    \midrule
    
    \multirow{3}{*}{ELECTRA\textsubscript{Large}} & SQuAD & 94.4 & 85.0 & 89.0 \\
    & SQuAD + AdversarialQA & 94.7 & \textbf{86.1} & \textbf{89.9} \\
    & SQuAD + AdversarialQA + SynQA & \textbf{94.8} & 85.7 & 89.2 \\
    
    \bottomrule
\end{tabular}
\caption{
Word-overlap \fone{} results for BERT, RoBERTa, and ELECTRA on the \squadone{} dev set and the AddSent and AddOneSent adversarial evaluation sets~\cite{jia2017adversarial}.
}
\label{tab:results_addsent}
\end{table*}
Table~\ref{tab:results_addsent} shows adversarial robustness performance evaluated on the AddSent and AddOneSent evaluation datasets introduced by~\citet{jia2017adversarial}.
We observe that even when trained only on \squadone{}, ELECTRA performs considerably better than RoBERTa in this setting, suggesting that it is substantially more robust ``out of the box''.
%
%%%%%%%%%%%%%%%%%%%%%%%%%%%%%%

%%%%%%%%%%%%%%%%%%%%%%%%%%%%%%
\section{Computational Resources}
\label{appendix:computational}

All experiments were run on single NVIDIA Tesla P100 GPUs. Models were trained for up to 14 epochs each taking approximately 2 hours to complete training.
Best model checkpoints and hyper-parameters were tuned for each experimental setting. The final model selected for each setting was based on validation performance across the \squad{} and AdversarialQA development sets.
The time taken for evaluation of the final models on each of the AdversarialQA test sets and the MRQA datasets was dependent on the number of examples.
%
%%%%%%%%%%%%%%%%%%%%%%%%%%%%%%

\end{document}